\documentclass{article}

\usepackage{PRIMEarxiv}

\usepackage[utf8]{inputenc} 
\usepackage[T1]{fontenc}    
\usepackage{url}            
\usepackage{booktabs}       
\usepackage{amsfonts}       
\usepackage{nicefrac}       
\usepackage{microtype}      
\usepackage{lipsum}
\usepackage{fancyhdr}       
\usepackage{graphicx}       
\usepackage[hidelinks]{hyperref}
\usepackage{orcidlink}
\graphicspath{{media/}}     
\usepackage{amsmath}
\usepackage{tikz}
\usetikzlibrary{arrows.meta, positioning, shapes.geometric}

\title{An Inpainting-Infused Pipeline for Attire and Background Replacement
}

\author{
  Felipe Rodrigues Perche Mahlow \orcidlink{0000-0001-9816-1440}\\
  Alana AI Research\\
  São Paulo, Brazil\\
  \texttt{felipe.mahlow@alana.ai} \\
  \And
  André Felipe Zanella \orcidlink{0000-0001-8828-6629}\\
  Alana AI Research\\
  São Paulo, Brazil\\
  \texttt{andre.zanella@alana.ai} \\
   \And
   William Alberto Cruz Castañeda \orcidlink{0000-0002-9803-1387}\\
  Alana AI Research\\
  São Paulo, Brazil\\
  \texttt{william.cruz@alana.ai} \\
    \And
  Marcellus Amadeus \orcidlink{0009-0002-7777-2562}\\
  Alana AI Research \\
  São Paulo, Brazil\\
  \texttt{marcellus@alana.ai} \\
}

\begin{document}
\maketitle

\begin{abstract}
In recent years, groundbreaking advancements in Generative Artificial Intelligence (GenAI) have triggered a transformative paradigm shift, significantly influencing various domains. In this work, we specifically explore an integrated approach, leveraging advanced techniques in GenAI and computer vision emphasizing image manipulation. The methodology unfolds through several stages, including depth estimation, the creation of inpaint masks based on depth information, the generation and replacement of backgrounds utilizing Stable Diffusion in conjunction with Latent Consistency Models (LCMs), and the subsequent replacement of clothes and application of aesthetic changes through an inpainting pipeline. Experiments conducted in this study underscore the methodology's efficacy, highlighting its potential to produce visually captivating content. The convergence of these advanced techniques allows users to input photographs of individuals and manipulate them to modify clothing and background based on specific prompts without manually input inpainting masks, effectively placing the subjects within the vast landscape of creative imagination.
\end{abstract}


\section{Introduction}

The extraordinary advancement in Generative Artificial Intelligence (GenAI) has caused a transformative shift in our approach to complex tasks incorporating various modalities such as text, audio, video, and image generation. GenAI, as a broad category, excels at creating synthetic data that can closely mimic real-world phenomena, showcasing its prowess in diverse creative applications. In text generation, models like OpenAI's GPT (Generative Pre-trained Transformer) \cite{openai2023gpt4} are revolutionizing how society writes. These models, trained on massive corpora of text data, exhibit an impressive ability to understand context, generate coherent paragraphs, and even complete sentences in a very consistent way \cite{roumeliotis2023chatgpt}. The ability to produce fluent and relevant textual content has established applications in natural language processing, content creation, and even automated writing \cite{huang2023role}. Audio generation models, exemplified by technologies such as Tacotron \cite{wang2017tacotron} and WaveNet \cite{oord2016wavenet}, have significantly advanced our ability to synthesize realistic speech patterns. These models take advantage of deep neural networks to capture the intricacies of human speech, producing natural-sounding voices and musical compositions with nuanced variations in tone, pitch, and rhythm \cite{ning2019review}.

Image generation, a focal point of our discussion, has witnessed the evolution of models such as DALL-E \cite{betker2023improving, ramesh2021zeroshot}, MidJourney \cite{midjourney}, and Stable Diffusion \cite{rombach2022high}, which can generate diverse and intricate images from textual prompts. Additionally, Generative Adversarial Networks (GANs) \cite{goodfellow2016deep} have been pivotal in generating high-fidelity, diverse images with applications ranging from art creation to realistic face synthesis, with achievements in computer vision, image processing, and multi-modal tasks \cite{wang2018perceptual, liu2019multistage, qiao2019mirrorgan}. The versatility of generative models across these modalities underscores their impact on reshaping the landscape of AI applications. Their ability to generate content indistinguishable from human-created artifacts has opened avenues for creative exploration, automation, and the development of novel applications across various industries, but also has been a cause of concern for the brief future \cite{walczak2023challenges}. 

Text-to-image (TTI) models are a research field devoted to methods and algorithm development tailored for written text into visual image conversion. A significant advancement in this area is the emergence and adoption of Latent Diffusion Models (LDMs) \cite{rombach2022high}, built upon the foundational framework of Diffusion Probabilistic Models (DPMs) \cite{ho2020denoising}. Diffusion models utilize a series of stochastic transformations applied sequentially to the probability distribution of an image \cite{croitoru2023diffusion}. This iterative process obtains detailed and realistic pictures while providing greater control over the generation process. While LDMs have shown remarkable results, their iterative sampling process can be computationally intensive, leading to slow generation. The emergence of Latent Consistency Models (LCMs) \cite{luo2023latent}  leverage Consistency Models \cite{song2023consistency} and directly predict the solution of augmented probability flow ODEs in latent space. This innovative approach reduces the need for numerous iterations, enabling rapid, high-fidelity image synthesis with minimal steps. 
 
An application within TTI models is Image Inpainting \cite{bertalmio2000image}. Inpainting involves the meticulous task of intelligently filling gaps or mending damaged regions in an image to achieve coherence and aesthetic appeal. Leveraging trained Diffusion Models has unlocked the potential for high-quality inpainting, effectively surmounting limitations inherent in traditional approaches \cite{lugmayr2022repaint}. However, in order to operate this technique, it is necessary to create \textit{masks} that indicate the regions of the picture that one wants to keep and the parts to alter. This process can be costly and laborious, assuming it deals with many images. Hence, this research strategically converges these two technological advances by proposing an innovative pipeline. This pipeline integrates a depth detection algorithm to automatically create inpainting masks for changing a person's clothing in a photograph. Simultaneously, we employed Stable Diffusion with Latent Consistency Models (LCMs) to generate new images for the background that we replaced using masks. To achieve this, we harness the advanced Intel MiDaS algorithm (Monocular Depth Estimation) \cite{ranftl2021vision, ranftl2020robust}, optimizing the understanding of spatial relationships in the image. Precise depth detection contributes to the creation of refined inpainting masks, enhancing the realism and adaptability of overlaying elements.

The paper is structured as follows: Section \ref{related_work}, we provide an overview of related work in the image synthesis field, especially Inpainting and Virtual Try-On methods. Section \ref{methods} delves into the methodology of our proposed approach for clothing and background exchange. The process encompasses depth estimation facilitated by the MiDaS algorithm, creation of inpainting masks based on depth information, generation, and replacement of backgrounds using Latent Consistency Models (LCMs), and the final steps of inpainting clothes and applying aesthetic enhancements. In Section \ref{experiments}, we present the experimental exploration of our methodology. We begin by detailing the hardware specifications used for the inference pipeline, followed by a pipeline results examination. These results demonstrate the adaptability and versatility of the proposed approach in generating diverse and visually appealing compositions in different scenarios. Section \ref{conclusion} concludes the paper by summarizing key findings, highlighting the effectiveness of the pipeline in image synthesis, and discussing potential avenues for future improvements. 

\section{Related Work}
\label{related_work}

\textbf{Text-to-image diffusion models}. Recently, there has been a significant evolution in text-to-image synthesis, particularly driven by the emergence of latent diffusion models (LDMs) \cite{rombach2022high}. Stable Diffusion \cite{rombach2022high} stands out as a notable work of LDMs, where the diffusion process occurs in the latent space rather than the original pixel space. Text embeddings from a pre-trained CLIP \cite{clip} model are fed into a UNet architecture enhanced with cross-attention. The recent Stable Diffusion XL \cite{sdxl} extends the original work with a more robust UNet and an additional and bigger CLIP text encoder, providing better text-guided conditioning. Other recent text-to-image model, DALL-E 3 \cite{betker2023improving}, achieves results as good as Stable Diffusion XL. This model is also a latent diffusion model trained with improved synthetic captions. Re-captioning the dataset using an image captioner has shown better CLIP Score performance than raw captions. In the other hand, other diffusion models, such as DeepFloyd-IF \cite{floyd}, Imagen \cite{imagen}, and eDiff-I \cite{ediff}, use a frozen T5 transformer \cite{t5} to extract text embeddings to be fed into the UNet, instead of CLIP text embeddings. 

Besides the advances in large text-to-image models, techniques for subject-driven text-to-image generation, like DreamBooth \cite{dreambooth} and Textual Inversion \cite{textualinversion} has emerged as suitable approach to generate personalized images of a subject with limited set of images. Our recent work demonstrates that these techniques, can generate high-quality images even when the training images are scarce and in poor-quality \cite{pampas}.

\textbf{Text-Guided Image Inpainting} One of the primary applications of text-to-image diffusion models is for image manipulation tasks. Applications for task-guided inpainting have been developed alongside latent diffusion models \cite{rombach2022high}. SmartBrush \cite{smartbrush} trains a diffusion model conditioned on an image, text, and a binary mask indicating the region the model should modify while preserving the image background. Blended Diffusion \cite{blended} utilizes a CLIP model to guide the output by calculating the CLIP Score between the image $+$ mask embedding and the text embedding. This value is incorporated into a sampling process of a pre-trained unconditional diffusion model, eliminating the need for training the model. Our method is similar since we eliminate the need for training but do not change the sampling process of diffusion models like DDIM \cite{ddim} and DDPM \cite{ddpm} by using an existing depth estimation algorithm while preserving the original image background.

\section{Methods}
\label{methods}
The following section unveils the pipeline. Commencing with depth estimation facilitated by the MiDaS algorithm \cite{ranftl2020robust}, this approach provides critical insights into the three-dimensional structure of scenes, subsequently guiding the creation of masks for inpainting purposes. The pipeline integrates a novel background generated through Stable Diffusion with Latent Consistency Models (LCMs). With an inpainting process, diverse clothing styles can be synthesized under prompt influence, offering creative control over the generated content.

\subsection{Depth Estimation and Inpainting Masks Creation}
\begin{figure*}
\centering
\includegraphics[width=\textwidth]{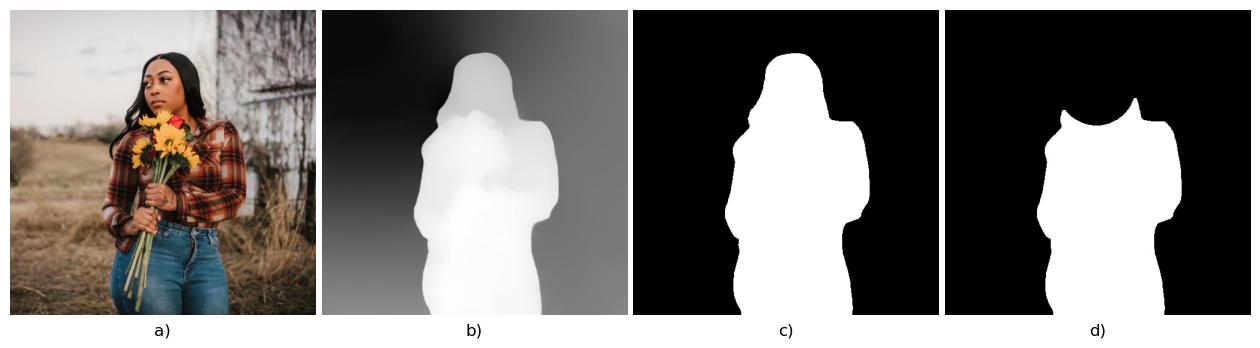}
\caption{Steps for inpainting mask processing. Image a) contains the original image of a woman holding a bouquet. Image b) is processed in black and white using the MiDaS algorithm. Image c) is the result after applying the threshold, and finally, image d) is the final mask after applying the facial recognition algorithm.}
\label{depth_plot}
\end{figure*}

The MiDaS algorithm is a network designed to predict depth maps from 2D images captured by a single camera. Developed by Intel and ETH Zurich researchers, MiDaS proposes a loss function to enable multi-objective learning data from different sources and explore pre-trained encoders. Once the depth map is obtained using MiDaS, it serves as a component in creating inpainting masks. In Figure \ref{depth_plot}, $a)$ is the original image, followed by $b$ the image converted to grey-scale after being processed by the MiDaS algorithm.

Inpainting masks are binary images that delineate the regions to be inpainted, typically corresponding to the background or undesired elements in the scene. Inpainting masks created in this work establish a threshold on the depth values. Pixels with depth values beyond a certain threshold are considered part of the subject, included in the inpainting mask as $1$ (white), and those who stayed below that threshold are classified as $0$, the background. Formally, the mask $M$ is defined as:

\begin{equation}
    M_i =
\begin{cases}
1, & \text{if } D_i > \text{Threshold} \\
0, & \text{otherwise}
\end{cases}
\end{equation}

Here, $D_i$ represents the depth value at pixel $i$, and the threshold is a user-defined parameter based on the specific requirements of the inpainting task, such as the person's distance on the frame or the presence of other objects on the image. In figure \ref{depth_plot} $c)$, we can see the resulting image after the established threshold, where the white pixels are those to be changed by the inpainting pipeline.

Face detection is a crucial step in our pipeline since we want to ensure that certain aspects, such as facial features, remain unaltered. In this step, the Haar Cascade Classifier \cite{viola2004robust} generates a classifier capable of identifying patterns associated with the target object, in this case, faces. The classifier is trained with the AdaBoost algorithm \cite{freund1999short}. We've applied the algorithm through the openCV library \cite{bradski2000opencv}. Before implementing face detection, is pre-processed the input image to enhance the algorithm's performance. The color image is converted to grayscale, as face detection often relies on variations in intensity rather than color information. The classifier works by sliding a window of various scales over the image, identifying regions that match the learned patterns of a face. Then, the mask creation involves generating elliptical masks centered on the detected face regions to indicate where we want to preserve the original characteristics. The size of the ellipse is determined by the bounding box dimensions, with a slight expansion to ensure complete coverage. 

Morphological operations ensure masks refine and smooth transitions between masked and inpainted regions. Opening operations with a suitable kernel reduces the noise and irregularities in the mask, resulting in a more accurate representation of the face region. The final inpainting mask, represented as a binary image, is obtained after applying all the above steps. In figure \ref{depth_plot} $d)$, we can see that the pixels corresponding to the face are now marked black, ensuring this way that they won't be affected by the inpainting algorithm.

\subsection{Background Generation and Replacement}

Integrating a new background into the image involves two key steps: Generating the new image and replacing the previous one. This section outlines the methodology for this operation.

Background generation is essential in achieving high-quality results since it establishes the main subject context, influencing the overall visual coherence and aesthetic appeal. Our work employs Latent Consistency Models (LCM) for the efficient generation of image backgrounds, utilizing the Hugging Face pipeline \footnote{https://huggingface.co/docs/diffusers/main/en/using-diffusers/lcm}. According to the authors, LCMs are distilled from any pre-trained Stable Diffusion (SD) model in approximately 4,000 training steps, equivalent to around 32 hours of A100 GPU computation. This process allows the generation of high-quality images in 2-4 steps, significantly accelerating image generation from text \cite{luo2023latent}. To conduct image generation from text, we utilize the StableDiffusionXLPipeline \cite{sdxl}, with the modified UNet distilled from the SDXL UNet using the LCM framework and the LCMScheduler.

The background replacement of the image using the depth map plays a fundamental role in achieving high-quality results. To perform this task, we employ the mask depicted in Figure \ref{depth_plot} c), where white pixels are preserved without changes, while black pixels replace corresponding pixels from the new background. This strategy aims to maintain consistency in highlighted regions of the image, preserving essential elements such as contours and distinctive features. It is crucial to emphasize that executing this background replacement precedes the application of inpainting techniques, as in our tests the algorithm exhibits enhanced harmonization capabilities in generating clothing details when presented with a specific background context in the image. This prior approach optimizes the general editing process, contributing to a more effective integration between the main object and the new visual environment.

\subsection{Generating the Clothes with the Inpainting Pipeline}

The clothing generation stage implements the SD-XL Inpainting 0.1 model \footnote{https://huggingface.co/diffusers/stable-diffusion-xl-1.0-inpainting-0.1}. The SD-XL Inpainting 0.1 was initialized with the weights of the stable-diffusion-xl-base-1.0 model, with a resolution of 1024x1024 \cite{von_Platen_Diffusers_State-of-the-art_diffusion}. Masks generation followed the procedure depicted in Figure \ref{depth_plot} d). In this context, the mask is crucial to guide the inpainting process, indicating the regions that will be preserved or modified in the final image. This approach allows for a precise replacement of clothing, ensuring that the face of the person in question remains unchanged.

During the inpainting process, several essential parameters guide and control the generation of the clothes. The text prompt is a guiding element that influences the style or content of the synthesized clothes. The original image and the mask image form the canvas for the inpainting operation. The \textit{guidance\_scale} parameter refers to the level of influence of the prompt applied to the original image. The \textit{num\_inference\_steps} parameter dictates the number of iterations involved, and the \textit{strength} parameter regulates the initial noise level applied to the image \footnote{https://github.com/CompVis/stable-diffusion/blob/main/README.md}. Lastly, the torch generator, denoted as \textit{generator}, is employed for manual seed control during the inpainting process, allowing the reproducibility of generated results. Together, these parameters collectively govern the inpainting process \cite{rombach2021highresolution}.

\subsection{Complete Pipeline Overview}

The complete pipeline involves a series of well-defined steps. Figure \ref{flowchart} illustrates the overall flowchart of the pipeline, highlighting the essential stages from start to finish.

\tikzset{
  >={Latex[width=2mm,length=2mm]},
  base/.style = {rectangle, rounded corners, draw=black,
                 minimum width=3.5cm, minimum height=1.2cm,
                 text centered, font=\sffamily, align=center},
  activityStarts/.style = {base, fill=green!30},
  startstop/.style = {base, fill=red!30},
  activityRuns/.style = {base, fill=green!30},
  process/.style = {base, minimum width=2.5cm, fill=orange!15,
                    font=\ttfamily},
}

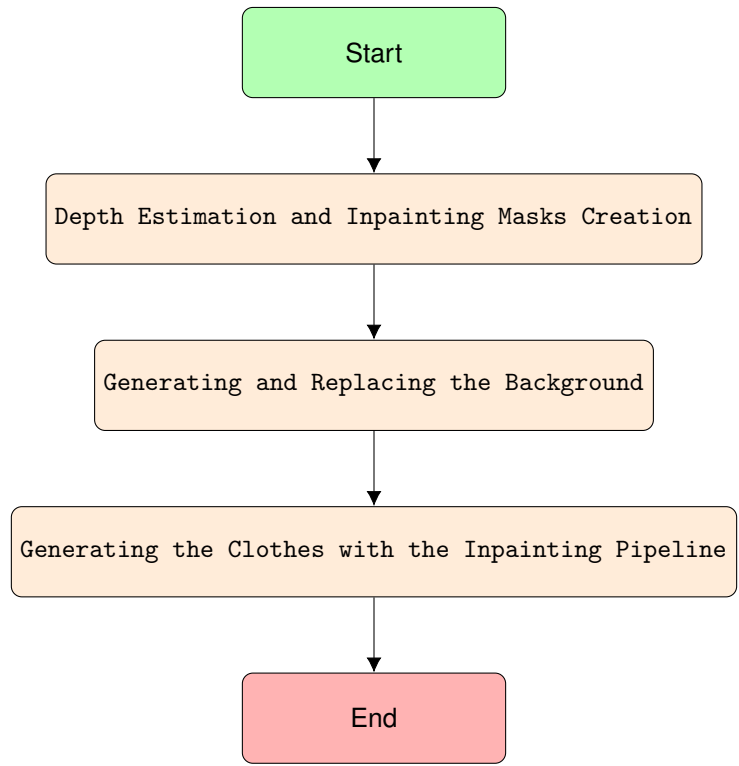
\begin{figure*}
\centering
\begin{tikzpicture}[node distance=1.0cm, every node/.style={fill=white, font=\sffamily}, align=center]

  \node (start) [activityStarts] {Start};
  \node (depth) [process, below=of start] {Depth Estimation and Inpainting Masks Creation};
  \node (background) [process, below=of depth] {Generating and Replacing the Background};
  \node (clothes) [process, below=of background] {Generating the Clothes with the Inpainting Pipeline};
  \node (end) [startstop, below=of clothes] {End};

  \draw[->] (start) -- (depth);
  \draw[->] (depth) -- (background);
  \draw[->] (background) -- (clothes);
  \draw[->] (clothes) -- (end);
\end{tikzpicture}

\caption{Flowchart of the complete pipeline}
\label{flowchart}
\end{figure*}

\begin{enumerate}
    \item Depth Estimation and Mask Creation: Utilize MiDaS algorithm for depth estimation, creating masks by thresholding depth values. Refine masks with face detection and morphological 
    operations. 
    \item Background Generation and Replacement: Generate a new background using Latent Consistency Models (LCMs). Replace the old background using the depth map to ensure consistency in highlighted regions.
    \item Clothing Synthesis with Inpainting: Employ SD-XL Inpainting 0.1 model guided by a text prompt. Use inpainting masks to control the synthesis process, preserving specific regions. Adjust parameters for style and content.    
\end{enumerate}

\section{Experiments}
\label{experiments}

In this section, we delve into the experimental exploration of our proposed methodology for clothing and background exchange. The aim is to provide the pipeline's capabilities and performance in various scenarios. We initiate our investigation by outlining the hardware specifications that underpin the computational processes. Subsequently, we present detailed insights into the results obtained from the pipeline, shedding light on its adaptability, versatility, and the nuances observed in different experimental contexts.

\subsection{Hardware Specifications}

The hardware configuration chosen for this study focuses on an inference pipeline without the need to train the model. Key components include the NVIDIA A100 PCIe GPU with 80GB of memory, selected for its parallel processing capabilities suitable for AI workloads.

\subsection{Pipeline Results}

\begin{figure*}
\centering
\includegraphics[width=\textwidth]{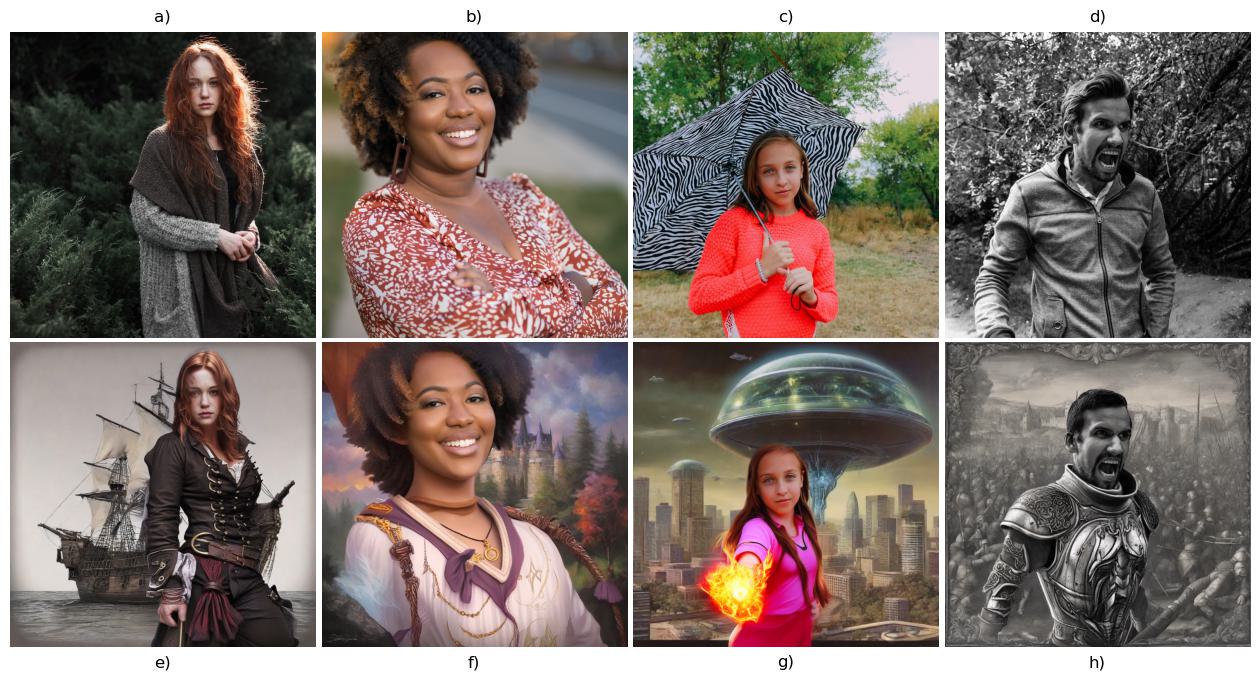}
\caption{Output using the pipeline for base images a)-d), and obtaining results e)-h). The prompts used for the background ($B_P$), clothes ($C_P$), and depth threshold $T_h$ were, respectively: e) pirate ship, pirate clothes, 0.6; f) magic castle painting, girl wizard clothes, 0.7; g) alien invasion, hero girl with fire powers, 0.6; h) black and white medieval battlefield painting, black and white metal battle armor, 0.5.}
\label{results_1}
\end{figure*}

\begin{figure*}
\centering
\includegraphics[width=\textwidth]{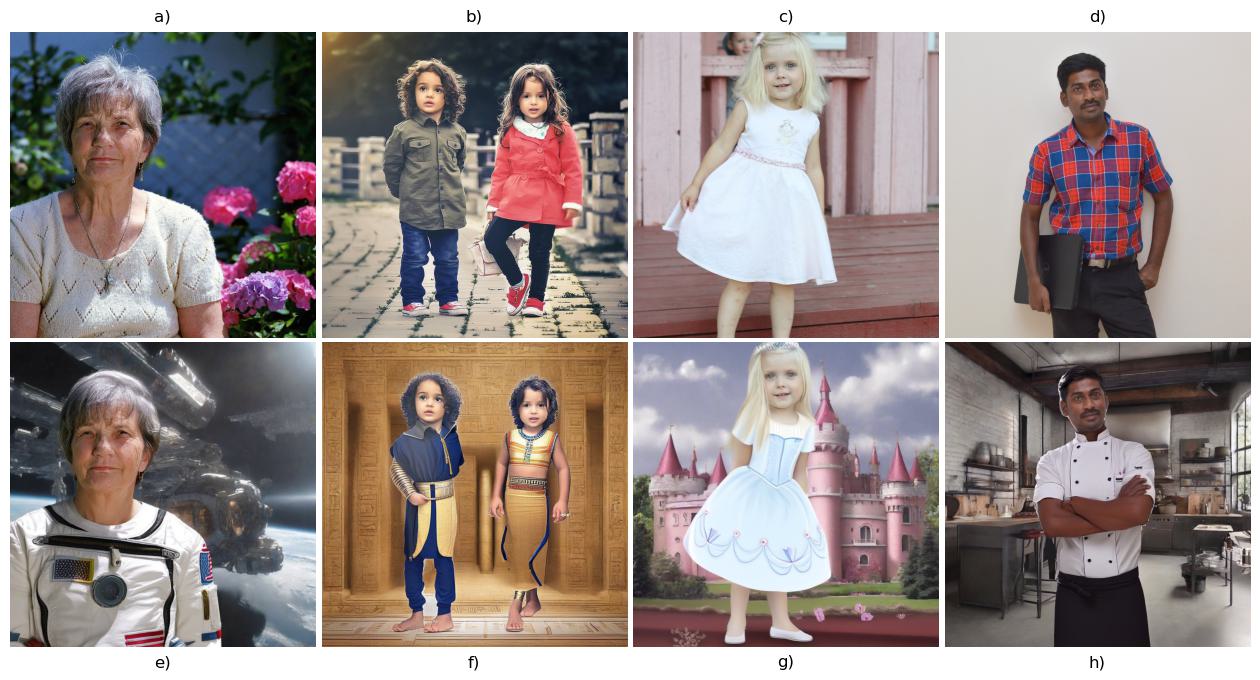}
\caption{Output using the pipeline for base images a)-d), and obtaining results e)-h). The prompts used for the background ($B_P$), clothes ($C_P$), and depth threshold $T_h$ were, respectively: e) space station exterior ultra HD 4K, astronaut suit ultra HD 4K, 0.7; f) ancient Egypt, ancient Egypt clothes pharaoh clothes, 0.7; g) princess castle, princess clothes, 0.6; h) industrial kitchen, chef clothes, 0.5.}
\label{results_2}
\end{figure*}

The pipeline results demonstrate the successful integration of the proposed methodology in clothing/background exchange based on specified prompts (found in the image captions). In the top row of Figures \ref{results_1} and \ref{results_2}, we showcase the original images to be modified, while the bottom rows display the outcomes of the modifications. It is essential to note that, due to subject identification relying on depth analysis, the pipeline operates effectively only when the subject is not influenced or blended with background elements.

In Figure \ref{results_1}, the pipeline demonstrates its adaptability by generating diverse scenes based on input images and corresponding prompts. Subjects of different ages, skin tones, and genders, set against various background configurations, were utilized to showcase the pipeline's ability to adapt to several scenarios, yielding excellent results. Figure \ref{results_1} $e)$, a notable algorithmic feature, highlights the tendency to harmonize clothing with the presented background. This phenomenon is observed in the inpainting pipeline, influenced by the order of image generation. For instance, if the generated background results in a cartoonish characteristic, the clothes may appear less realistic, resembling drawings or paintings rather than photographs.

In Figure \ref{results_1} $f)$, another characteristic of the inpainting pipeline is depicted—a propensity to "slim down" individuals during image generation. Related to the challenge of generating loosely dispersed hair, especially afro-textured hair, it is crucial to be attentive to these aspects to avoid reinforcing stereotypes and to ensure inclusivity and representation for all. Figure \ref{results_1} $g)$ demonstrates the flexibility to adjust the threshold to highlight objects close to the body, exemplified by the presented umbrella case. Lastly, Figure \ref{results_1} $h)$ emphasizes the capability to introduce color elements through prompts. All these results are illustrated in the presented case, showcasing the option to utilize the black-and-white spectrum.

Similarly, Figure \ref{results_2} underscores the pipeline's versatility across different contexts. Base images combined with specific prompts yield unique and contextually relevant outputs. The seamless fusion of background themes and clothing details highlights the pipeline's capability to generate visually coherent and engaging compositions. In Figure \ref{results_2}e), particularly when compared to Figure \ref{results_2}f, we observe that despite a significant age difference, both scenarios perform well. Notably, Figure \ref{results_2}f the effectiveness of incorporating more than one person in the same image is highlighted. Achieving optimal results is easier when using a single person per image. Additionally, Figure \ref{results_2}f showcases the pipeline's successful representation of full-body subjects. It is fundamental to mention some classic challenges in text-to-image generation models, such as generating hands, feet, and arm positions in specific circumstances.

Figure \ref{results_2}g) reaffirms the tendency to slim down individuals and cartoonized clothing based on the generated background. Figure \ref{results_2}h presents a more "realistic" situation, demonstrating the capability to restructure arm positions, remove previously existing objects, and the increased ease of repositioning people against white backgrounds. Throughout the pipeline's development, we observed that the depth detection algorithm's accuracy was higher when using photos with a blurred background, such as the portrait mode on smartphones. Collectively, these results highlight the effectiveness of the proposed pipeline for image synthesis in generating contextually relevant and visually appealing compositions based on input images and textual prompts. Diverse background elements integration, clothing details, and depth considerations contribute to the system's ability to produce creative outputs.

\section{Conclusion}
\label{conclusion}

The presented work explores an integrated approach to image manipulation, leveraging advanced techniques in generative artificial intelligence and computer vision. The developed pipeline encompasses various stages, including inpainting, background replacement, and facial feature preservation. The integration of depth estimation with the MiDaS algorithm facilitated the creation of precise inpainting masks that contribute to generating realistic and contextual accurate results. The inpainting process using Latent Diffusion Models (LDMs) proves effective in seamlessly modifying the people's clothes in the images.

Background replacement is achieved with the density estimation to identify the replaced pixels. Stable Diffusion XL (SDXL) and Latent Consistency Models (LCMs) generate background contexts quickly and effectively. These features open for creative applications, from personalized photo editing to the production of visually impactful content for diverse mediums. The proposed pipeline addresses the limitations of traditional inpainting methods by incorporating state-of-the-art generative models and depth-aware techniques. 

With the continuous progress of the technology, future work in this domain could explore additional improvements, such as sophisticated generative components integration, fine-tuning of parameters for specific applications, and the exploration of novel techniques for inpainting and enhancement. In conclusion, generative AI and computer vision techniques open new possibilities for image manipulation, paving the way for visually stunning creation and contextually rich visual content.

\section*{Acknowledgments}
We acknowledge the support of the National Council for Scientific and Technological Development (CNPq) and Alana AI for funding the research.

\section*{Funding}
This research is supported by the National Council for Scientific and Technological Development (CNPq/MCTI/SEMPI Number 021/2021 RHAE - 383035/2022-8), Ministry of Science, Technology and Innovation, and Alana AI. 

\section*{Contributions}
FM and AZ conducted the experiments. WC and MS provided oversight for the research. All authors contributed to the paper's composition, with MS and WC contributing to the study's conception. The final manuscript was reviewed and approved by all authors.

\section*{Interests}
The authors declare that they have no competing interests.

\section*{Materials}
The datasets generated and/or analyzed during the current study are available on request. All the images used in where selected from Pixabay \footnote{https://pixabay.com/service/license-summary/} database and are royalty-free under their Content License Summary. 

\bibliographystyle{apalike-sol}
\bibliography{main}

\end{document}